\documentclass{article}

\usepackage{microtype}
\usepackage{graphicx}
\usepackage{booktabs} 

\usepackage{hyperref}

\usepackage[accepted]{icml2024}

\usepackage{amsmath}
\usepackage{amssymb}
\usepackage{mathtools}
\usepackage{amsthm}

\usepackage[capitalize,noabbrev]{cleveref}

\theoremstyle{plain}

\theoremstyle{definition}

\theoremstyle{remark}

\usepackage[textsize=tiny]{todonotes}

\usepackage{svg}
\usepackage{array}
\usepackage{multirow}
\usepackage{caption}
\usepackage{subcaption}
\usepackage{adjustbox}
\usepackage{titlesec}

\DeclarePairedDelimiter\ceil{\lceil}{\rceil}
\DeclarePairedDelimiter\floor{\lfloor}{\rfloor}
\titleformat{\subsubsection}{\normalfont\large\bfseries}{\thesection}{1em}{}

\begin{document}

\twocolumn[
    \icmltitle{MultiResFormer: Transformer with Adaptive Multi-Resolution Modeling for General Time Series Forecasting}
    
    
    
    \icmlsetsymbol{equal}{*}
    
    \begin{icmlauthorlist}
    \icmlauthor{Linfeng Du}{l6,uoft}
    \icmlauthor{Ji Xin}{l6}
    \icmlauthor{Alex Labach}{l6}
    \icmlauthor{Saba Zuberi}{l6}
    \icmlauthor{Maksims Volkovs}{l6}
    \icmlauthor{Rahul G. Krishnan}{uoft,vector}
    \end{icmlauthorlist}
    
    \icmlaffiliation{l6}{Layer 6 AI}
    \icmlaffiliation{uoft}{University of Toronto}
    \icmlaffiliation{vector}{The Vector Institute}
    
    \icmlcorrespondingauthor{Linfeng Du}{linfeng@cs.toronto.edu}
    
    
    \vskip 0.5in
]


\printAffiliationsAndNotice{}  

\begin{abstract}
Transformer-based models have greatly pushed the boundaries of time series forecasting recently. State-of-the-art methods encode time series by segments called \textit{patches} using one or a fixed set of patch lengths and strides. While different patching schemes enable modeling time series under various time scales, i.e., resolutions, methods that rely on pre-defined time scales could result in a lack of ability to capture the variety of intricate temporal dependencies present in real-world multi-periodic series. In this paper, we propose MultiResFormer to model time series with adaptive time scales. Concretely, each model block operates on multiple branches with different patch lengths, which are chosen as detected salient periodicities of times series data. A shared Transformer block is applied to jointly model intraperiod and interperiod dependencies under all the time scales. We conduct extensive evaluations on long- and short-term forecasting datasets comparing MultiResFormer with state-of-the-art baselines. MultiResFormer outperforms patch-based Transformer baselines on long-term forecasting tasks and also consistently outperforms CNN baselines by a large margin, while using fewer parameters than other methods.
\end{abstract}

\section{Introduction}
Time series forecasting is a common sequence modeling problem and is crucial to a wide range of real-world applications including power distribution, weather forecasting, traffic flow scheduling, and disease propagation analysis. Accurately forecasting into a long horizon is of immense importance for risk aversion and decision making in these fields, and so there is great interest in applying deep learning models \cite{Lim2020Time}, as they have shown to be capable of learning compact representations that are generalizable to a range of time series analysis tasks.

Motivated by the superior performance of Transformer-based models demonstrated across various domains \cite{Kalyan2021AMMUS,Khan2022Transformers,Karita2019Comparative}, time series Transformers have received an abundance of research attention in the past few years \cite{Wen2023Transformers}. Progress has been made to adapt the vanilla Transformer \cite{Vaswani2017Attention} to tackle unique challenges for time series forecasting. Modifications include 1) proposing efficient attention mechanisms for sub-quadratic attention computation \cite{Li2019Enhancing,Zhou2021Informer,Liu2022Pyraformer}; 2) breaking the point-wise nature of dot-product attention for segment or series level dependency modeling \cite{Li2019Enhancing,Wu2021Autoformer,Zhou2022FEDformer,Nie2023Time,Zhang2023Crossformer}; 3) modeling sequences at multiple time scales with hierarchical representation learning \cite{Li2019Enhancing,Cirstea2022Triformer,Shabani2023Scaleformer,Anonymous2024Multi}.

Time series Transformers with multi-scale modeling use modified input schemes or architectures to explicitly capture meaningful patterns across different time scales. Existing methods leverage pre-defined hierarchies of fine-to-coarse resolutions formed by down-sampling with fixed rates \cite{Liu2022Pyraformer,Cirstea2022Triformer,Shabani2023Scaleformer}. A more recent work seek to form multiple resolutions by different patching schemes \cite{Anonymous2024Multi}. Such pre-defined resolutions could introduce redundancy and be missing of important scales. Therefore, they may not generalize well to a wide range of series with different variation patterns and periodicities, where correlated intraperiod and interperiod variations co-exist at each scale.

\begin{figure}
    \centering
    \includegraphics[width=\linewidth]{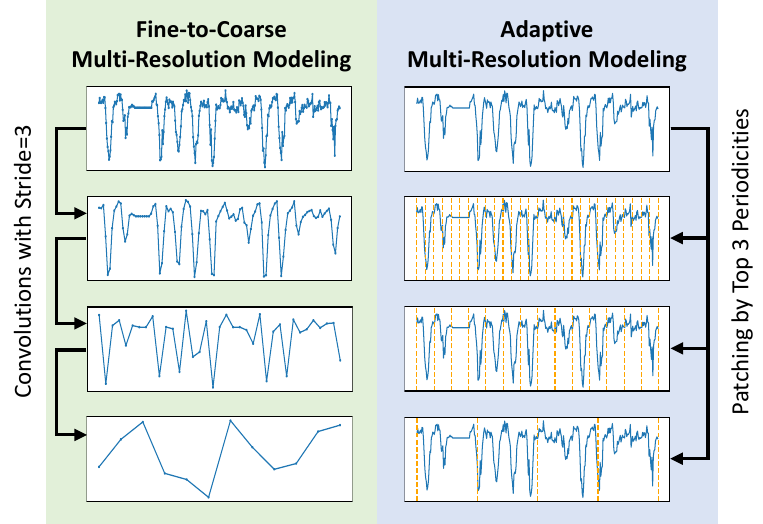}
    \caption{Different ways to form multiple resolutions. \textbf{Left}: the hierarchy formed by iterative down-sampling with fixed rates. Each resolution is a coarsened version of the previous one. \textbf{Right}: Multiple resolutions adaptively formed by patching with salient periodicities. This breaks the dependency between resolutions and preserves all information in the original series at each resolution.}
    \label{fig:resolution}
\end{figure}

We propose that neural networks for time series should first identify relevant time-scales in data, and use this information to guide downstream prediction of trajectories. To this end, we develop new Transformer blocks that use adaptive multi-resolution modeling and use them as the building blocks within an encoder network. 
Rather than using a hierarchy of fixed fine-to-coarse resolutions, our method first identifies underlying periodicities in the data to create a semantically useful multi-resolution view; see \Cref{fig:resolution}.

By composing multiple novel Transformer blocks that enable \emph{adaptive} multi-resolution modeling, we obtain the MultiResFormer. Each block contains multiple branches that operate at different resolutions. The inputs to the branches are formed by non-overlapping patching where the patch lengths are set in accordance to the detected salient periodicities. We repurpose the two core Transformer sublayers, namely Multi-headed attention (MHA) and position-wise feed-forward network (FFN), for modeling interperiod and intraperiod variations. Specifically, we leverage MHA for interperiod variation modeling for its capability to model global patch-wise dependency in $\mathcal{O}(1)$ computations, and FFN for intraperiod variation modeling as MLPs are capable of modeling even more complicated dependencies within the whole series \cite{Zeng2023Are,Chen2023TSMixer}.

Adaptivity does not come for free and our work proposes several key design decisions to realize a practical neural network architecture.

First, a critical problem is how to enable parameter-sharing across different resolution branches; doing so enables the model to scale to an arbitrary number of resolutions with a fixed number of parameters, as well as to share learned characteristics across different scales. This could help prevent overfitting and make learning easier. While not knowing the patch lengths (periodicities) in advance forbids the usage of linear projection layers to embed the patches, simply padding the patches to a same length would harm the dependency modeling process as MHA and FFN cannot mask out redundant feature dimensions. We instead develop an interpolation scheme which up- or down-samples the patches to the same length. We further add a resolution embedding to each branch input to enable scale-awareness.

Second, it is crucial for modeling efficiency to reduce the complexity associated with completing a forward pass in the model. MultiResFormer's encoder-only architecture with in-block patching design removes the need for an embedding layer by generating representations that match the input size. This embedding-free design draws motivation from the recent success of very simple linear models \cite{Zeng2023Are}, and greatly reduces the parameter burden of the prediction head compared to recent patch-based time series Transformer \cite{Nie2023Time}, where patches are embedded into a high-dimensional space.

To sum up, the key contributions of this work are as follows:

\begin{itemize}
    \item[1.] We propose MultiResFormer, a Transformer-based model which performs efficient in-block adaptive multi-resolution modeling using periodicity-guided patching. We leverage key Transformer sublayers to achieve effective interperiod and intraperiod variation modeling.
    \item[2.] We propose an interpolation scheme to foster parameter sharing across different resolutions, and an additive resolution embedding to enhance scale awareness.
    \item[3.] We conduct extensive experiments over long- and short-term forecasting benchmark datasets to demonstrate the effectiveness of our proposed method against strong baselines. In addition, we conduct thorough ablation studies to empirically justify our design choices.
\end{itemize}

\section{Related Work}
\subsection{Time Series Transformers}
\label{TST}
Transformer-based models have achieved great success in various domains including natural language processing \cite{Kalyan2021AMMUS}, computer vision \cite{Khan2022Transformers}, healthcare \citep{labach2023duett}, and audio signal processing \cite{Karita2019Comparative}. This is largely attributed to the global dependency modeling capacity of the attention mechanism as well as ensembling across multiple attention heads.

A limitation of leveraging attention mechanisms for modeling longer series is their quadratic complexity w.r.t. the sequence length. One line of research aims for sub-quadratic attention computation by limiting the number of query-key pairs to be computed. To achieve $\mathcal{O}(N\log N)$ complexity, LogTrans \cite{Li2019Enhancing} uses sparse attention blocks where each token attends to others with an exponential step size. Informer \cite{Zhou2021Informer} instead proposes an entropy-based measurement to filter out uninformative keys. For $\mathcal{O}(N)$ complexity, Triformer \cite{Cirstea2022Triformer} and Pyraformer \cite{Liu2022Pyraformer} adopt CNN-like approaches where a sequence of local attention operations and temporal pooling is applied. Linear complexity is achieved by limiting the depth of the pyramidal structure. Meanwhile, patch-based methods \cite{Nie2023Time,Zhang2023Crossformer} quadratically reduce attention's complexity since the patches are sampled with stride, resulting in $\mathcal{O}(N^2/S^2)$ complexity where $S$ is the stride. In MultiResFormer blocks, the patch lengths are decided according to detected salient periodicities and the strides are set equal to the patch lengths, which enables a reduction in computational complexity while retaining information relevant to forecasting.

An important characteristic of time series data is that values within each individual time step do not convey much information \cite{Nie2023Time}. \cite{Li2019Enhancing} points out that when operating on time step-based tokens, canonical dot-product attention may overlook localized patterns and be misguided by the similarity between two individual time steps. Current research focuses on breaking this limitation by modeling relationships across segments or series rather than individual time steps. LogTrans \cite{Li2019Enhancing} first proposes to use convolution instead of point-wise MLPs to generate queries and keys so as to enhance the locality of attention mechanism. Autoformer \cite{Wu2021Autoformer} and FEDformer \cite{Zhou2022FEDformer} model period-based dependencies by aggregating lagged series with normalized auto-correlation scores (Autoformer) or computing attention under Fourier basis (FEDformer). More recent works draw the idea of patching from vision \cite{Dosovitskiy2021Image,Liu2021Swin} and propose patch-based modeling for time series data \cite{Nie2023Time,Zhang2023Crossformer}. MultiResFormer follows this approach, but instead of relying on fixed patch length and stride, we adaptively select patch lengths to better fit the varied multi-periodic nature of real-world time series data. Compared to Autoformer and FEDformer which only focus on interperiod variations, MultiResFormer also models intraperiod variations with the FFN sublayer.

For multivariate time series, most Transformer-based methods \cite{Li2019Enhancing,Zhou2021Informer,Wu2021Autoformer,Zhou2022FEDformer,Cirstea2022Triformer,Liu2022Pyraformer} apply a convolution layer to project the variates into fix-sized embeddings. PatchTST \cite{Nie2023Time} finds empirically that such a channel-mixing embedding approach causes quick over-fitting when training on real-world multivariate datasets. This finding motivates PatchTST's channel-independent modeling where cross-variate dependency is only implicitly captured via batching. Meanwhile, a concurrent work \cite{Zhang2023Crossformer} explicitly leverages attention layers to capture cross-variate dependencies, yet fails to achieve better performance than PatchTST. Given this context, MultiResFormer follows PatchTST and adopts a channel-independent modeling paradigm.

\subsection{Multi-Resolution Time Series Modeling}
\label{MRTSM}

Multi-resolution time series modeling lies at the core of temporal convolution-based methods, as they operate on a fine-to-coarse hierarchy formed by down-sampling with fixed rates. As each layer exponentially increases the receptive field, a stack of $\mathcal{O}(\log N)$ layers is required to capture the dependency between every input pair, resulting in $\mathcal{O}(N\log N)$ complexity in total. SCINet \cite{Minhao2022SCINet} proposes sample convolution where the input series is further divided by even and odd indices at each convolution layer to enlarge the receptive field by a constant factor. MICN \cite{Wang2023MICN} proposes a two-staged approach to achieve $\mathcal{O}(N)$ complexity. The series is first down-sampled by a strided convolution with large kernel and stride, then an isomorphic convolution layer is applied to model global dependencies for the down-sampled series. Drawing inspirations from these methods, recent work in Transformer-based models rely on similar hierarchies to reduce the quadratic complexity of vanilla dot-product attention \cite{Cirstea2022Triformer,Liu2022Pyraformer} and perform multi-resolution modeling \cite{Shabani2023Scaleformer}.

A concurrent work \cite{Anonymous2024Multi} utilizes pre-defined patch lengths and models both inter-patch and intra-patch dependencies via separate attention mechanisms. For intra-patch dependency, attention is computed pair-wise between time steps within patches. MultiResFormer adaptively decide the patch lengths on-the-fly based on the time series frequencies and use the FFN layer to model intra-patch dependency. TimesNet \cite{Wu2023TimesNet} leverage adaptive multi-resolution modeling albeit in a different style of neural architecture. There, the 1D time series is reshaped into 2D matrices according to the detected salient periodicities. Treating variates as channels, a 2D convolutional backbone (e.g. an Inception block \cite{Szegedy2015Going}) is applied to jointly model intraperiod and interperiod variations. This strategy has several drawbacks. Firstly, only a single 2D convolution layer could be employed at each model block since it requires the processed matrices to remain the same shape so as to recover the input length via flattening. This limitation in modeling longer-range dependencies cannot be fundamentally addressed by ensembling multiple small-sized kernels (as in Inception networks) which brings additional sequential operations. Secondly, the 2D variation modeling paradigm inevitably imposes channel-mixing which has been identified to be prone to over-fitting \cite{Nie2023Time}.

\begin{figure*}[t!]
    \centering
    \begin{subfigure}[b]{0.45\textwidth}
        \includegraphics[width=\textwidth]{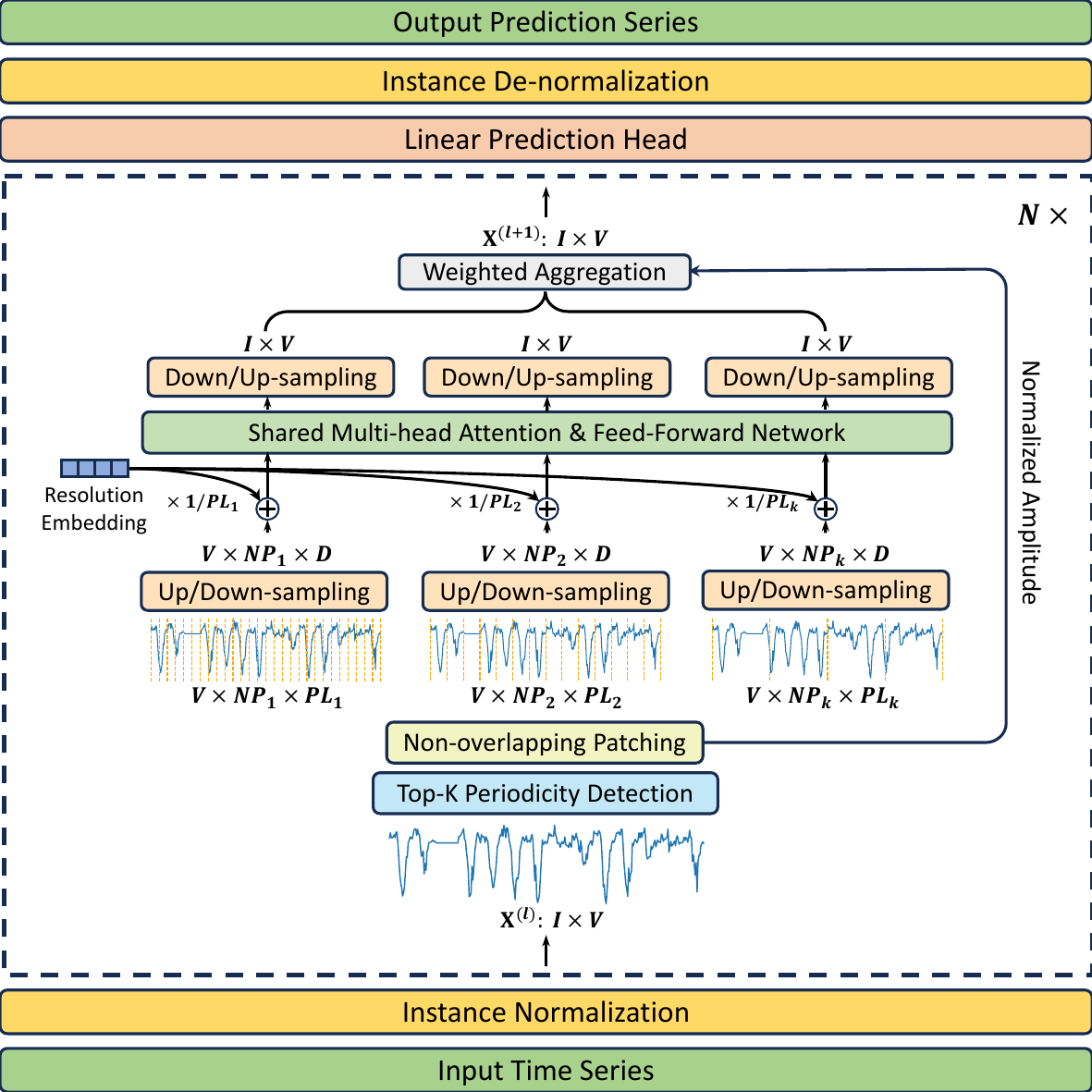}
    \end{subfigure}
    \centering
    \begin{subfigure}[b]{0.45\textwidth}
        \includegraphics[width=\textwidth]{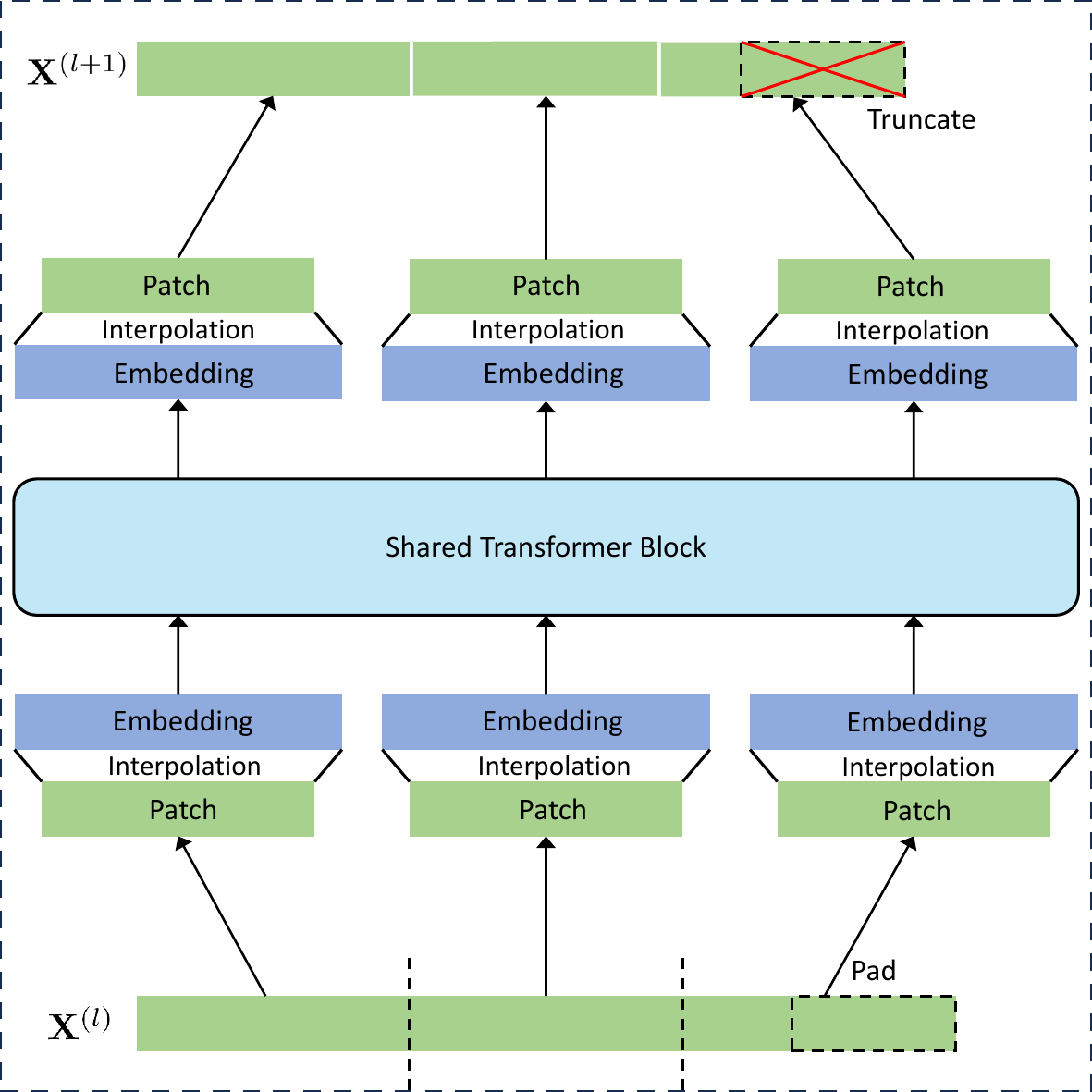}
    \end{subfigure}
    \caption{MultiResFormer architecture. \textbf{Left}: the MultiResFormer model consists of Instance Normalization/De-normalization at the beginning/end; in between there is a stack of $N$ MultiResFormer blocks (in the dotted box), followed by a Linear Prediction Head. We mark the shape of the intermediate tensors for better understanding. $PL_i$ denotes the patch length used in the $i$-th resolution branch, which corresponds to the $i$-th salient periodicity of $\mathbf{X}^{(l)}$. $NP_i$ denotes the number of patches after non-overlapping patching. $D$ denotes the model size that we interpolate each of the patches to. \textbf{Right}: at each resolution branch, $\mathbf{X}^{(l)}$ is padded, split into patches, interpolated into fixed-size embeddings, passed through a shared Transformer Block, interpolated back to its original patch length, flattened, and truncated.}
    \label{fig:model}
\end{figure*}

\section{Adaptive Multi-Resolution Time Series Modeling with Transformers}

We consider the problem of time series forecasting. Given a collection of observations within a look-back window of size $I$: $\mathbf{X}_{1...I} = (\mathbf{x}_1,...,\mathbf{x}_I) \in \mathbb{R}^{I \times V}$ where the value at each time step $\mathbf{x}_t \in \mathbb{R}^V$ is a scalar for univariate series ($V = 1$) or a vector for multivariate series ($V \ge 2$), our goal is to forecast the values of $O$ future time steps $\mathbf{X}_{I+1...I+O} = (\mathbf{x}_{I+1},...,\mathbf{x}_{I+O}) \in \mathbb{R}^{O \times V}$.
We describe the neural network of the MultiResFomer hierarchically. The left half of \Cref{fig:model} shows the overall architecture of our model. 

\subsection{MultiResFormer}
The MultiResFormer model consists of a stack of $N$ MultiResFormer blocks, a linear prediction head, and a normalization--denormalization layer pair.
At the beginning, we apply instance normalization to the input series so that the input to the subsequent MultiResFormer blocks has zero-mean and unit-variance along the temporal dimension; at the end, instance de-normalization is applied to restore the first two moments of the look-back window back to the prediction series. These coupled operations, referred to as Reversible Instance Normalization (RevIN) \cite{Kim2022Reversible}, are also applied in previous work \cite{Nie2023Time,Wu2023TimesNet} to tackle temporal distribution shift which is common in real-world time series. In between RevIN, $N$ MultiResFormer blocks are applied to produce a representation of the input series, followed by a linear prediction head which projects the output of the final MultiResFormer block to produce the prediction of the future time steps (i.e., from $\mathbb{R}^{I \times V}$ to $\mathbb{R}^{O \times V}$). Note that the prediction head is shared across variates to learn cross-variate characteristics and to reduce the number of parameters.

Prior time series Transformers either generate a high-dimensional representation for each time step \cite{Zhou2021Informer,Wu2021Autoformer,Zhou2022FEDformer} or each patch of time steps \cite{Nie2023Time,Zhang2023Crossformer}, so that the encoder output may have a different shape than the input. 
However, MultiResFormer ensures that the encoder output has the same shape $I \times V$ as the input series. This is achieved using the in-block patching and aggregation design and has the additional benefit of significantly reducing the parameter overhead of the linear prediction head (compared to the state-of-the-art patch-based PatchTST model \cite{Nie2023Time}). As we empirically demonstrate in \Cref{sec:efficiency}, MultiResFormer achieves a strong trade-off between performance and number of parameters.

The remainder of this section will describe the details of the MultiResFormer block. The $l$-th MultiResFormer block takes series $\mathbf{X}^{(l-1)} \in \mathbb{R}^{I \times V}$ as input where $\mathbf{X}^{(0)} = \mathbf{X}_{1...I}$ is the original input time series. It has two components. 

\begin{itemize}
    \item The periodicity-aware patching module has $k$ resolution branches and it performs patching based on the detected salient periodicities (\Cref{sec:SPD}).
    \item The Transformer Encoder block is shared across all resolution branches to capture patterns across different resolutions (\Cref{sec:MRMSTB}). 
\end{itemize} 

We then aggregate the representations derived within each resolution branch into the block output $\mathbf{X}^{(l)} \in \mathbf{R}^{I \times V}$ (\Cref{sec:AA}). In the remainder of this section, we omit the layer index superscript $l$ for brevity.

\begin{table*}[t]
\centering
\caption{Statistics of long-term and short-term forecasting datasets.}
\adjustbox{max width=\linewidth}
{
    \begin{tabular}{c|cccccccc|cccccccc}
         \toprule
         Dataset & ETTh1 & ETTh2 & ETTm1 & ETTm2 & Weather & Electricity & Traffic & ILI & \multicolumn{6}{c}{M4} \\
         \midrule
         Variates & 7 & 7 & 7 & 7 & 21 & 321 & 862 & 7 & 1 & 1 & 1 & 1 & 1 & 1 \\
         Frequency & Hourly & Hourly & 15min & 15min & 10min & Hourly & Hourly & Weekly & Yearly & Quarterly & Monthly & Weekly & Daily & Hourly \\
         Instances & 17420 & 17420 & 69680 & 69680 & 52696 & 26304 & 17544 & 966 & 23000 & 24000 & 48000 & 359 & 4227 & 414 \\
         Domain & Electricity & Electricity & Electricity & Electricity & Weather & Electricity & Transportation & Illness & Demographic & Finance & Industry & Macro & Micro & Other \\
         \bottomrule
    \end{tabular}
}
\label{tab:dataset}
\end{table*}

\subsection{Salient Periodicity Detection}
\label{sec:SPD}

Instead of relying on a pre-defined hierarchy of resolutions, each MultiResFormer block constructs multiple resolution branches based on the salient periodicites of the input series.

Specifically, we leverage the Fast Fourier Transform (FFT) to decompose the input $\mathbf{X}$ of each MultiResFormer block into Fourier basis and to identify the top \(k\) salient periodicities based on the amplitude of each frequency:
\begin{equation}
    \begin{aligned}
        \mathbf{A} &= \operatorname{Avg}(\operatorname{Amp}(\operatorname{FFT}(\mathbf{X}))), \\
        \{f_1,...,f_k\} &= \underset{f_* \in \{1,...,\floor{\frac{I}{2}}\}}{\operatorname{ argTopk}}(\mathbf{A}), \\
        {\rm Period}_i &= \ceil*{\frac{I}{f_i}}.
    \end{aligned}    
\end{equation}
Here, FFT is performed for each variate independently. The \(\operatorname{Avg}(\cdot)\) operation averages the computed amplitudes along the variate dimension to generate a unified set of periodicities for all the variates, so as to parallelize the computations within each resolution branch and reflect overall data characteristics. For batch implementation in practice, we further average the amplitudes across different examples. Empirically we do not find this averaging to severely affect model performance, as the periodicities across different variates and examples are relatively stable. Since we only use the detected salient periodicities as patch lengths in the next step, differentiability is not required for the periodicity detection method. Therefore, any automated periodicity detection technique including auto-correlation based methods or hybrid approaches \cite{Puech2020Fully} could be applied here.

\begin{table*}[ht]
    \caption{Multivariate long-term forecasting results with MultiResFormer. The prediction lengths are set to $O \in \{24,36,48,60\}$ for ILI and $O \in \{96,192,336,720\}$ for the other datasets. The results are averaged across the 4 prediction lengths. The best results are in \textbf{bold} and the second best are \underline{underlined}. See \Cref{tab:main_LTF} in Appendix for full results.}
    \centering
    \adjustbox{max width=\linewidth}{
    \begin{tabular}{c|c|cc|cc|cc|cc|cc|cc|cc|cc}
        \toprule
        \multicolumn{2}{c|}{Models} & \multicolumn{2}{c|}{MultiResFormer} & \multicolumn{2}{c|}{PatchTST} & \multicolumn{2}{c|}{TimesNet} & \multicolumn{2}{c|}{MICN} & \multicolumn{2}{c|}{DLinear} & \multicolumn{2}{c|}{FEDformer} & \multicolumn{2}{c|}{Autoformer} & \multicolumn{2}{c}{Informer} \\
        \multicolumn{2}{c|}{Metric} & MSE & MAE & MSE & MAE & MSE & MAE & MSE & MAE & MSE & MAE & MSE & MAE & MSE & MAE & MSE & MAE \\
        \midrule
        \multicolumn{2}{c|}{ETTh1}
        & \textbf{0.411} & \textbf{0.424} & \underline{0.420} & \underline{0.431} & 0.467 & 0.459 & 0.510 & 0.508 & 0.436 & 0.446 & 0.431 & 0.455 & 0.470 & 0.473 & 1.061 & 0.815 \\
        \midrule
        \multicolumn{2}{c|}{ETTh2}
        & \textbf{0.342} & \textbf{0.384} & \underline{0.348} & \underline{0.390} & 0.407 & 0.421 & 0.573 & 0.523 & 0.434 & 0.450 & 0.440 & 0.449 & 0.452 & 0.465 & 4.470 & 1.727 \\
        \midrule
        \multicolumn{2}{c|}{ETTm1}
        & \textbf{0.353} & \textbf{0.381} & \underline{0.355} & 0.383 & 0.406 & 0.415 & 0.391 & 0.416 & 0.360 & \underline{0.382} & 0.417 & 0.441 & 0.536 & 0.496 & 0.903 & 0.716 \\
        \midrule
        \multicolumn{2}{c|}{ETTm2}
        & \textbf{0.256} & \textbf{0.314} & \underline{0.257} & \underline{0.316} & 0.295 & 0.332 & 0.307 & 0.362 & 0.276 & 0.340 & 0.298 & 0.349 & 0.321 & 0.363 & 1.547 & 0.881 \\
        \midrule
        \multicolumn{2}{c|}{Weather}
        & \textbf{0.227} & \textbf{0.262} & \underline{0.228} & \underline{0.263} & 0.257 & 0.284 & 0.251 & 0.308 & 0.248 & 0.302 & 0.307 & 0.358 & 0.345 & 0.386 & 0.662 & 0.559 \\
        \midrule
        \multicolumn{2}{c|}{Electricity}
        & \underline{0.166} & \underline{0.258} & \textbf{0.164} & \textbf{0.257} & 0.195 & 0.296 & 0.188 & 0.295 & 0.167 & 0.265 & 0.215 & 0.327 & 0.256 & 0.357 & 0.384 & 0.455 \\
        \midrule
        \multicolumn{2}{c|}{Traffic}
        & \underline{0.409} & \underline{0.275} & \textbf{0.395} & \textbf{0.264} & 0.629 & 0.338 & 0.536 & 0.312 & 0.435 & 0.297 & 0.613 & 0.380 & 0.657 & 0.411 & 1.015 & 0.568 \\
        \midrule
        \multicolumn{2}{c|}{ILI}
        & \textbf{1.866} & \textbf{0.883} & \underline{2.080} & \underline{0.961} & 2.261 & 0.928 & 2.764 & 1.135 & 2.294 & 1.072 & 3.353 & 1.283 & 3.462 & 1.298 & 5.223 & 1.566 \\
        \bottomrule
    \end{tabular}}
    \label{tab:main_LTF_short}
\end{table*}
\subsection{Multi-Resolution Modeling with a Shared Transformer Block}
\label{sec:MRMSTB}

We use the $k$ detected periodicities as patch lengths to segment the series in $k$ different ways, and apply a shared Transformer block to improve parameter efficiency and make the model size invariant to the choice of $k$. While methods that use a fixed patch length \cite{Nie2023Time,Zhang2023Crossformer} apply shared linear transformations to project the patches into high-dimensional embeddings, such patch embedding layers are not applicable here since we compute the patch lengths on-the-fly.

A naive way to derive fix-sized patches is to pad or truncate the segmented patches to the same length. However, the inconsistency in the semantic of the embedding dimensions would harm model performance since the MHA and FFN sublayers lack a mechanism to mask out uninformative (padded) dimensions. Instead, we leverage interpolation to up- or down-sample patches to the same length. As a common operation in time series analysis \cite{Wang2023MICN,Minhao2022SCINet}, interpolation effectively preserves the temporal characteristics within the original patches.


The patching and interpolation at each resolution branch is shown in the right half of \Cref{fig:model}, at the bottom. Given ${\rm Period}_i$ corresponding to one resolution branch, we split the input of the current block $\mathbf{X}$ into non-overlapping patches of length ${\rm Period}_i$. If the length of $\mathbf{X}$ is not divisible by ${\rm Period}_i$, paddings will be added by repeating the last value of $\mathbf{X}$.
To obtain fixed-sized patch representations, we linearly interpolate the patches to length $d$, which is a hyperparameter that also serves as the model size. The shape of the patch-based representation of the input series is then $V \times \ceil{\frac{I}{{\rm Period}_i}} \times d$.

Now that each resolution branch operates on same-sized patches, in order to better distinguish different resolutions, we propose to use a resolution embedding that consists of a learnable embedding scaled inversely proportional to the corresponding periodicity so as to incur smoothness over different resolution branches. This can be written as
\begin{equation}
    \begin{aligned}
        \textbf{\text{Res}}_i &= \textbf{\text{RE}} \times \frac{1}{{\rm Period}_i},
    \end{aligned}
\end{equation}
where $\mathbf{RE} \in \mathbb{R}^{d}$ is the randomly initialized learnable embedding shared across the entire model. The Transformer input for each patch is the sum of the interpolated patch representation and the corresponding resolution embedding, so that the subsequent Transformer block is aware of the original resolution (i.e., patch length).

The original patch lengths are set according to the salient periodicities and we adopt non-overlapping patching. This patching scheme naturally entails two types of variations, i.e., \textit{interperiod} and \textit{intraperiod} variations \cite{Wu2023TimesNet}, corresponding to the different dimensions of the patches. As we interpolate the patches to the same length, we leverage a shared Transformer block to jointly capture the two types of variations within each resolution branch in sequence.
Specifically, we leverage MHA to capture patch-wise dependencies, which corresponds to interperiod variation modeling, and we leverage FFN layers for capturing dependencies within each patch, which corresponds to intraperiod variation modeling. This is in line with the general use of Transformers for modelling a sequence of tokens, where MHA layers model global dependencies between tokens and FFN layers process tokens individually.




\newcolumntype{C}[1]{>{\centering\let\newline\\\arraybackslash\hspace{0pt}}m{#1}}
\def\arraystretch{1.3}
\begin{table*}[ht]
    \footnotesize
    \caption{Univariate short-term forecasting results with MultiResFormer. Results are averaged across several datasets with different sampling frequencies. See \Cref{tab:main_STF} for full results. The best results are in \textbf{bold} and the second best are \underline{underlined}. Baseline results are taken from \cite{Wu2023TimesNet}.}
    \centering
    \begin{adjustbox}{width=0.85\linewidth}
    \begin{tabular}{c|c|C{1.41cm}|C{1.41cm}|C{1.41cm}|C{1.41cm}|C{1.41cm}|C{1.41cm}|C{1.41cm}|C{1.41cm}}
        \toprule
        \multicolumn{2}{c|}{Models} & MultiRes. & TimesNet & N-HiTS & N-BEATS & DLinear & FED. & Auto. & In. \\
        \midrule
        \multicolumn{2}{c|}{SMAPE} & \textbf{7.617} & \underline{7.664} & 7.775 & 7.692 & 9.163 & 8.264 & 8.479 & 13.325 \\
        \multicolumn{2}{c|}{MASE} & \textbf{2.061} & \underline{2.103} & 2.106 & 2.135 & 2.948 & 2.174 & 2.367 & 6.730 \\
        \multicolumn{2}{c|}{OWA} & \textbf{0.885} & \underline{0.897} & 0.908 & 0.903 & 1.152 & 0.952 & 1.006 & 2.203 \\
        \bottomrule
    \end{tabular}
    \end{adjustbox}
    \label{tab:main_STF_short}
\end{table*}

\subsection{Adaptive Aggregation}
\label{sec:AA}

The representation from the $i$-th resolution branch is of shape $V \times \ceil{\frac{I}{{\rm Period}_i}} \times d$. Before aggregating the results from different branches, we need to recast them back to the same shape $I \times V$. We show the process in the right half of \Cref{fig:model} (top half). Here we interpolate the patches back to the original patch length ${\rm Period}_i$, and recover the series by flattening the last two dimensions. If extra paddings are added in the initial padding step, the corresponding part of the output will be truncated.



Finally, we collect the representations from all $k$ resolution branches, each with shape $I \times V$. We aggregate them and perform a weighted sum to produce the final output of this MultiResFormer block. The weights are based on normalized amplitudes of the corresponding frequencies:
\begin{equation}
    \begin{aligned}
        \{A_1,...,A_k\} &= \underset{f_* \in \{1,...,\floor{\frac{I}{2}}\}}{\operatorname{Topk}}(\mathbf{A}) \\
        \{w_1,...,w_k\} &= \operatorname{Softmax}(\{A_1,...,A_k\}) \\
    \end{aligned}
\end{equation}


\section{Experiments}

We evaluate the performance of MultiResFormer on a number of long-term and short-term time series forecasting datasets. The long-term forecasting datasets include four Electricity Transformer Temperature (ETT) datasets (ETTh1, ETTh2, ETTm1, ETTm2), plus Weather, Electricity, Traffic and Influenza-Like Illness (ILI) datasets. These datasets were gathered and pre-processed by \citet{Wu2021Autoformer} and have been widely used by recent work for benchmarking purposes. For short-term forecasting, we use the M4 datasets \cite{Makridakis2020M4} which includes 6 subsets corresponding to different sampling frequencies. The statistics of the two types of datasets are summarized in \Cref{tab:dataset}, and the implementation details are described in \Cref{ID}.

We compare MultiResFormer with state-of-the-art methods including PatchTST \cite{Nie2023Time}, TimesNet \cite{Wu2023TimesNet}, MICN \cite{Minhao2022SCINet}, DLinear \cite{Zeng2023Are}, FEDformer \cite{Zhou2022FEDformer}, Autoformer \cite{Wu2021Autoformer} and Informer \cite{Zhou2021Informer}. We do not include Transformer-based methods that leverage pre-defined hierarchies of resolutions like Pyraformer \cite{Liu2022Pyraformer} and Triformer \cite{Cirstea2022Triformer} since they have been shown in previous work \cite{Nie2023Time,Wu2023TimesNet} to underperform the above selected baselines. It is also worth mentioning that the more recent Pathformer \cite{Anonymous2024Multi} adopts a much shorter look-back window length of 96, as compared to 336 which is used by MultiResFormer and PatchTST. As a result, their reported performance is not as competitive. Since the codebase for this work is not yet available, we leave the comparison to future work.

\begin{table}[t]
    \caption{Ablation studies: multivariate long-term forecasting results on the 4 ETT datasets with input length $I=336$ and prediction length $O \in \{96,192,336,720\}$. The best results are in \textbf{bold} and the second best are \underline{underlined}.}
    \centering
    \begin{adjustbox}{width=\linewidth}
    \begin{tabular}{c|c|cc|cc|cc|cc}
        \toprule
        \multicolumn{2}{c|}{Models} & \multicolumn{2}{c|}{MultiResFormer} & \multicolumn{2}{c|}{w/o Res. Emb.} & \multicolumn{2}{c|}{PatchTST Ens.} & \multicolumn{2}{c}{PatchTST} \\
        \midrule
        \multicolumn{2}{c|}{Metric} & MSE & MAE & MSE & MAE & MSE & MAE & MSE & MAE \\
        \midrule
        \multirow{4}{*}{\rotatebox{90}{ETTh1}}
        & 96 & \textbf{0.367} & \textbf{0.391} & \underline{0.370} & \underline{0.394} & 0.374 & 0.396 & 0.376 & 0.399 \\
        & 192 & \textbf{0.405} & \textbf{0.414} & \underline{0.409} & \underline{0.416} & 0.413 & 0.419 & 0.411 & 0.419 \\
        & 336 & \textbf{0.430} & \textbf{0.432} & \underline{0.432} & \underline{0.432} & 0.434 & 0.432 & 0.439 & 0.438 \\
        & 720 & \underline{0.441} & \underline{0.461} & \textbf{0.441} & \textbf{0.460} & 0.442 & 0.460 & 0.456 & 0.467 \\
        \midrule
        \multirow{4}{*}{\rotatebox{90}{ETTh2}}
        & 96 & \textbf{0.272} & \textbf{0.333} & \underline{0.275} & \underline{0.336} & 0.288 & 0.344 & 0.277 & 0.339 \\
        & 192 & \textbf{0.338} & \textbf{0.375} & \underline{0.342} & \underline{0.380} & 0.355 & 0.384 & 0.344 & 0.382 \\
        & 336 & \textbf{0.364} & \textbf{0.400} & \underline{0.368} & \underline{0.401} & 0.370 & 0.401 & 0.377 & 0.407 \\
        & 720 & \underline{0.392} & \underline{0.428} & \textbf{0.392} & \textbf{0.427} & 0.399 & 0.431 & 0.393 & 0.431 \\
        \bottomrule
    \end{tabular}
    \end{adjustbox}
    \label{tab:ablation}
\end{table}

\subsection{Main Results}

For long-term forecasting, We follow the experiment settings of PatchTST \cite{Nie2023Time} where the lookback window length is set to 104 for ILI and 336 for other datasets. For other baselines, we reuse the scripts provided by \citet{Nie2023Time} which summarize the best hyperparameter settings of the existing baselines for reproducing the results.\footnote{For long-term forecasting, we rerun all baseline experiments due to an identified error in the data loader which drops the last batch during testing. This error is inherited by all baselines in their codebases and causes the test set to vary across methods using different batch sizes.} For short-term forecasting on M4, we follow the experiment settings of TimesNet \cite{Wu2023TimesNet}.

\Cref{tab:main_LTF} and \Cref{tab:main_STF} show the multivariate long-term forecasting and univariate short-term forecasting results, respectively. Compared with the CNN-based TimesNet which surpasses previous Transformer-based methods, MultiResFormer outperforms it by a significant margin (overall 16.5\% reduction on MSE and 10.7\% on MAE). Compared with the state-of-the-art baseline PatchTST, MultiResFormer also consistently outperforms it in most datasets and prediction lengths. \Cref{fig:periodicity} in the Appendix shows the statistics of the top 3 salient periodicities of the long-term forecasting datasets. The salient periodicity detection method effectively generates useful resolutions. For example, the daily scale is typically selected for electricity-related datasets with sample rate of both an hour (ETTh1, ETTh2) and 15 minutes (ETTm1, ETTm2), while weekly scale is selected for the Traffic dataset which exhibits clear differentiation in weekday and weekend patterns. Also, the stability of the detected periodicities could also help make learning easier.

\subsection{Multi-Resolution Design Choices}
One of the major contributions of our paper is the in-block multi-resolution patching design. Alternatively, a straightforward approach to enable multiple resolutions would be to ensemble multiple PatchTST models with different patch lengths. To show the importance of our design choices, we evaluate the performance of PatchTST ensembles and compare with MultiResFormer.

For PatchTST ensembles, we set patch lengths to the salient periodicities averaged across the training portion of the datasets (denoted as PatchTST Ens.). Since the patch lengths are computed in advance, we apply different patch embedding layers and a shared PatchTST backbone. In addition to PatchTST Ens., we also train a MultiResFormer model without the resolution embedding. The results are listed in \Cref{tab:ablation}. Ensembling PatchTST with different patch lengths does not always result in better performance, while our MultiResFormer with resolution embedding consistently outperforms PatchTST. This shows that although setting patch lengths based on detected periodicities is helpful for Transformer models, naively ensembling PatchTSTs is not sufficient to capture the extra information. The in-block patching mechanism along with resolution embeddings are critical to the final performance.

\subsection{Varying Look-back Window Size}

\begin{figure}[t]
    \centering
    \includegraphics[scale=0.21]{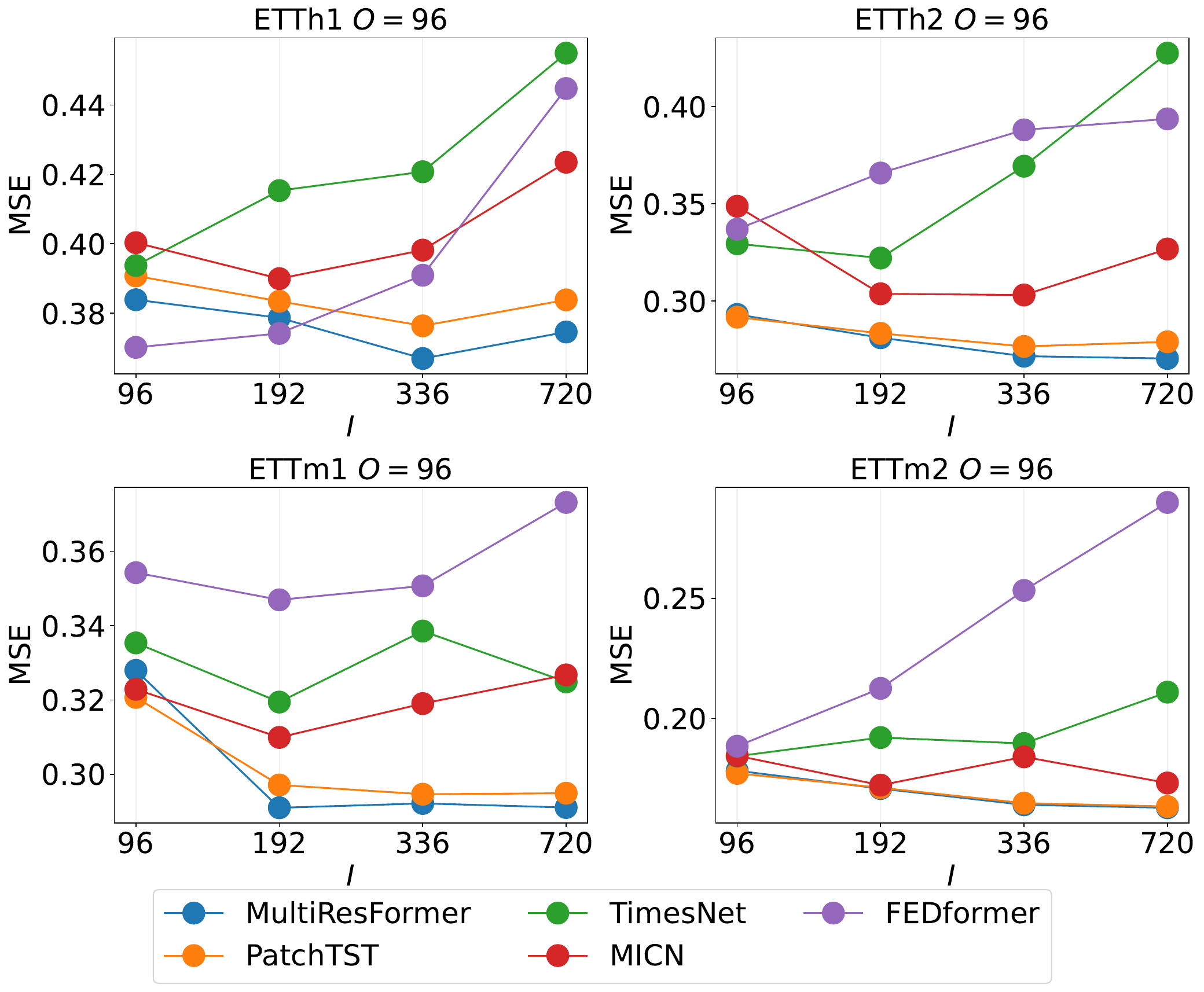}
    \caption{Forecasting performance with varying look-back windows on 4 ETT datasets.}
    \label{fig:lookback}
\end{figure}

The PatchTST paper \cite{Nie2023Time} claims that many time series forecasting models do not necessarily benefit from a longer look-back window. We conduct a similar analysis for MultiResFormer and its baselines in \Cref{fig:lookback}.

A larger look-back window provides more information of the past, which could potentially improve model performance. However, as we increase the look-back window length, we observe consistent boosting effects  only in PatchTST and our model MultiResFormer.
Previous work \cite{Zeng2023Are} has identified that earlier Transformer-based models including FEDformer, Autoformer and Informer tend to have worse performance as we further increase the look-back window length. \citet{Nie2023Time} attribute this effect to early-stage over-fitting caused by the application of channel-mixing embedding layers. Our observations for the Transformer-based models are consistent with these previous papers, and we notice similar phenomenon for CNN-based methods like TimesNet and MICN which also leverage such embedding layers. In contrast, MultiResFormer has the property of benefiting from longer look-back windows similar to PatchTST; furthermore, its improvement compared to PatchTST is not confined to specific look-back window sizes.

\subsection{Efficiency Comparison}
\label{sec:efficiency}
\begin{figure}[t]
    \centering
    \includegraphics[scale=0.25]{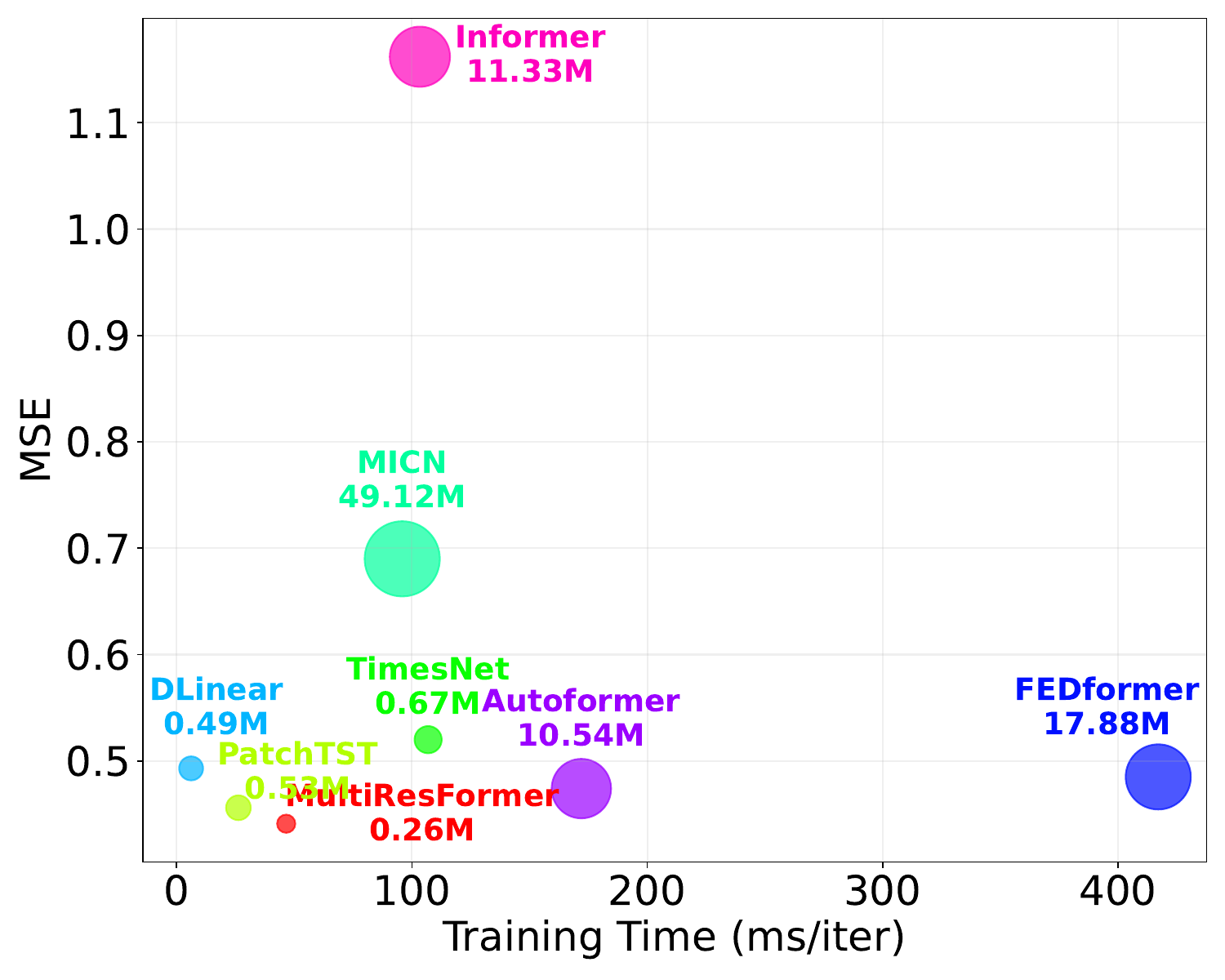}
    \caption{Model efficiency comparison on the ETTh1 dataset.}
    \label{fig:efficiency}
\end{figure}

\Cref{fig:efficiency} shows a comparison of model efficiency of MultiResFormer along with the compared baselines tested on the ETTh1 dataset. Despite having the lowest parameter count, MultiResFormer achieves the best MSE. Moreover, in terms of training speed, MultiResFormer is significantly faster than CNN-based TimesNet and comparable to PatchTST, since the sequential operation of modeling multiple resolution branches does not add much to training time.

\subsection{Representation Analysis}

\begin{figure}[t]
    \centering
    \includegraphics[scale=0.3]{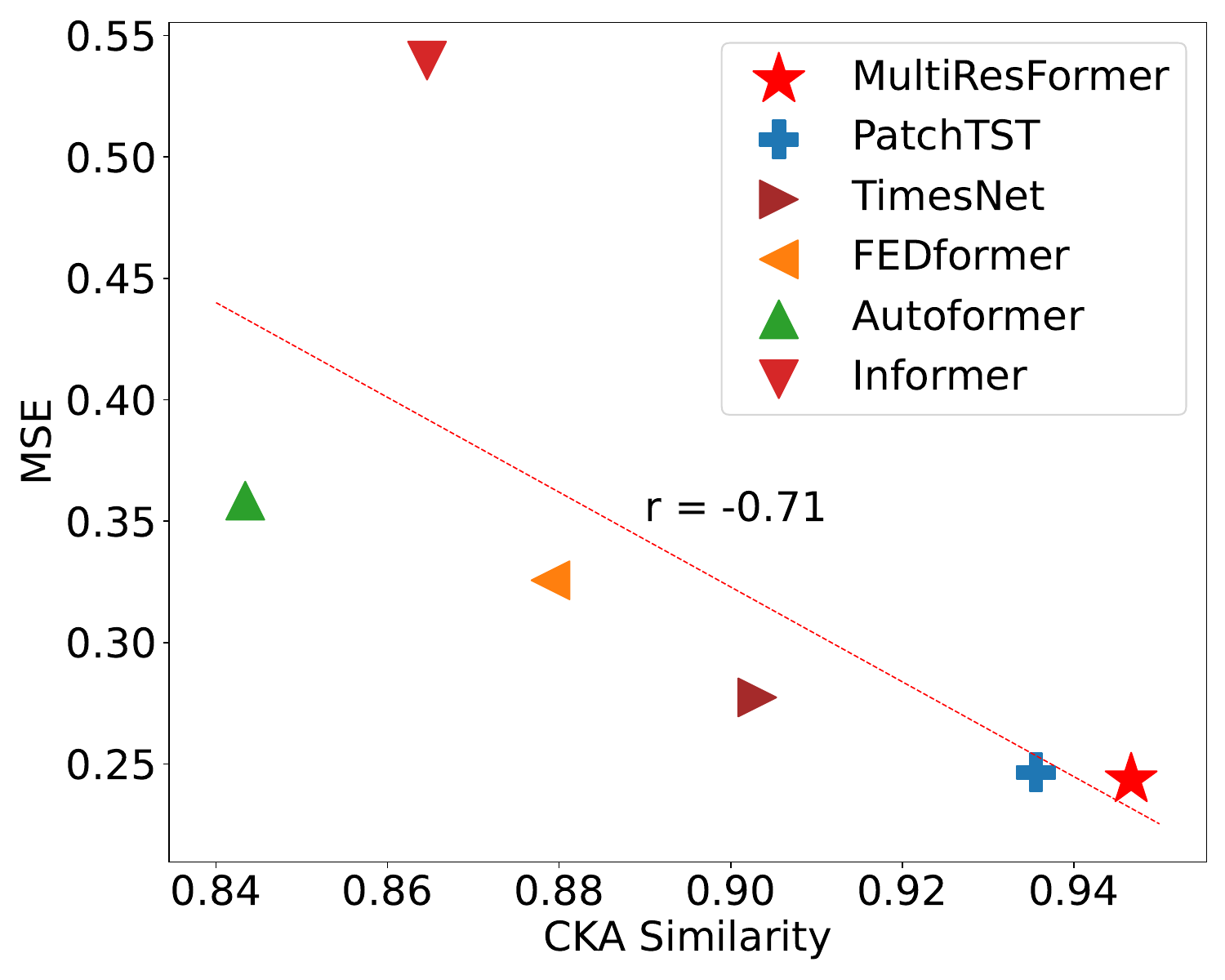}
    \caption{Representation analysis for long term forecasting on the Weather dataset.}
    \label{fig:cka}
\end{figure}

We aim to explain the better performance of MultiResFormer by studying the representations learned by different MultiResFormer blocks in terms of their centered kernel alignment (CKA) similarity \cite{Kornblith2019Similarity}. It is identified in previous work \cite{Wu2023TimesNet} that a higher CKA similarity between the first and last layer representations positively correlates to better model performance for low level time series analysis tasks like forecasting. Here we perform similar study on Weather dataset. \Cref{fig:cka} shows the average CKA similarity on the test set of Weather dataset obtained with various trained models. Among which, MultiResFormer has the lowest CKA similarity and the best forecasting performance, indicating that MultiResFormer learns more appropriate representations for this task.

\section{Conclusion}

This paper proposes a Transformer model capable of adaptively modeling time series under multiple resolutions. This is achieved by on-the-fly detection of underlying periodicities within each Transformer block to determine salient resolutions, with patched representations at each resolution being passed to a separate branch modeled by shared Transformer sublayers. The in-block patching design removes the need of an embedding layer so that the encoder generates representations with the same size as the input, greatly reducing the parameter burden of the final linear prediction head compared to similar models. Within each Transformer block we propose an interpolation scheme to allow parameter sharing across resolution branches and an resolution embedding to make the model resolution aware. Extensive experiments on long- and short-term forecasting benchmarks show the superior forecasting performance of our model in comparison to seven state-of-the art algorithms.

\section*{Impact Statement}

Our paper proposes a deep learning model for time series forecasting. Time series forecasting has various applications in the real world, for example, predicting future currency exchange rate (or other financially sensitive data), electricity consumption, traffic volume, and climate change. These use cases are partially reflected in the datasets we use and being able to accurately predict on these datasets will affect related areas. On the other hand, the scope of our paper is very similar to that of previous related work, and we find that the minor difference is unlikely to bring about noticeable social consequences.

\bibliography{icml2024}
\bibliographystyle{icml2024}

\newpage
\appendix
\onecolumn

\section{Implementation Details}
\label{ID}
Our model is trained using ADAM \cite{Kingma2015Adam} optimizer with a learning rate of $1e^{-4}$ for long-term forecasting experiments and $1e^{-3}$ for short-term forecasting experiments. The batch size is set to 32. Training is early stopped if the loss on the validation set does not decrease for ten epochs. All models are implemented in PyTorch \cite{Paszke2019PyTorch} and trained on NVIDIA V100 32GB GPUs. We set $d$ to the most frequent salient periodicity for each dataset and $k$ to 3. To ensure reproducibility, we will provide implementation code for our model along with the camera-ready version of the paper.

\section{Multi-Periodicity of Benchmark Time Series}

\begin{figure*}[ht]
    \centering
    \includegraphics[width=\linewidth]{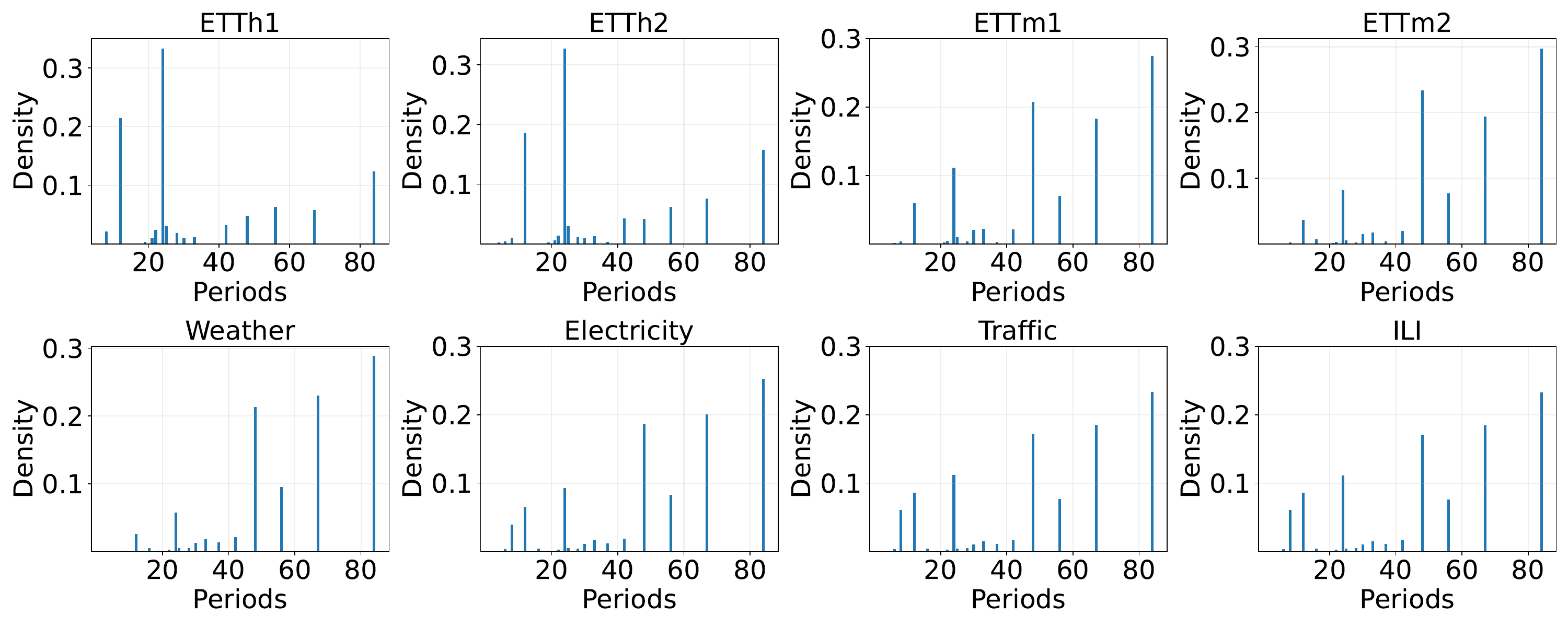}
    \caption{Statistics of the top 3 salient periodicities detected based on length-336 look-back windows across various long-term forecasting datasets. Each look-back window is transformed to the Fourier domain via FFT. We collect the top 3 frequencies with the highest amplitude and calculate their corresponding periodicity for each look-back window. The y axis shows the normalized count of each selected periodicity. Periodicities with higher density are more stable across different look-back windows.}
    \label{fig:periodicity}
\end{figure*}

\newpage
\section{Full Results}

\begin{table*}[ht]
    \caption{Multivariate long-term forecasting results with MultiResFormer. The prediction lengths are set to $O \in \{24,36,48,60\}$ for ILI and $O \in \{96,192,336,720\}$ for the other datasets. The best results are in \textbf{bold} and the second best are \underline{underlined}.}
    \centering
    \adjustbox{max width=\linewidth}{
    \begin{tabular}{c|c|cc|cc|cc|cc|cc|cc|cc|cc}
        \toprule
        \multicolumn{2}{c|}{Models} & \multicolumn{2}{c|}{MultiResFormer} & \multicolumn{2}{c|}{PatchTST} & \multicolumn{2}{c|}{TimesNet} & \multicolumn{2}{c|}{MICN} & \multicolumn{2}{c|}{DLinear} & \multicolumn{2}{c|}{FEDformer} & \multicolumn{2}{c|}{Autoformer} & \multicolumn{2}{c}{Informer} \\
        \multicolumn{2}{c|}{Metric} & MSE & MAE & MSE & MAE & MSE & MAE & MSE & MAE & MSE & MAE & MSE & MAE & MSE & MAE & MSE & MAE \\
        \midrule
        \multirow{4}{*}{\rotatebox{90}{ETTh1}}
        & 96 & \textbf{0.367} & \textbf{0.391} & 0.376 & \underline{0.399} & 0.394 & 0.416 & 0.400 & 0.428 & 0.377 & 0.401 & \underline{0.370} & 0.411 & 0.459 & 0.462 & 0.932 & 0.765 \\
        & 192 & \textbf{0.405} & \textbf{0.414} & \underline{0.411} & \underline{0.419} & 0.449 & 0.448 & 0.447 & 0.463 & \textbf{0.405} & \textbf{0.414} & 0.420 & 0.443 & 0.443 & 0.452 & 0.965 & 0.763 \\
        & 336 & \textbf{0.430} & \textbf{0.432} & \underline{0.439} & \underline{0.438} & 0.505 & 0.476 & 0.505 & 0.506 & 0.469 & 0.466 & 0.451 & 0.464 & 0.505 & 0.488 & 1.185 & 0.872 \\
        & 720 & \textbf{0.441} & \textbf{0.461} & \underline{0.456} & \underline{0.467} & 0.520 & 0.496 & 0.690 & 0.636 & 0.493 & 0.504 & 0.485 & 0.500 & 0.474 & 0.490 & 1.162 & 0.858 \\
        \midrule
        \multirow{4}{*}{\rotatebox{90}{ETTh2}}
        & 96 & \textbf{0.272} & \textbf{0.333} & \underline{0.277} & \underline{0.339} & 0.329 & 0.368 & 0.349 & 0.405 & 0.294 & 0.360 & 0.337 & 0.377 & 0.422 & 0.451 & 3.081 & 1.376 \\
        & 192 & \textbf{0.338} & \textbf{0.375} & \underline{0.344} & \underline{0.382} & 0.401 & 0.411 & 0.511 & 0.489 & 0.384 & 0.419 & 0.432 & 0.438 & 0.434 & 0.438 & 6.062 & 2.059 \\
        & 336 & \textbf{0.364} & \textbf{0.400} & \underline{0.377} & \underline{0.407} & 0.457 & 0.456 & 0.599 & 0.543 & 0.443 & 0.461 & 0.498 & 0.489 & 0.485 & 0.489 & 5.225 & 1.903 \\
        & 720 & \textbf{0.392} & \textbf{0.428} & \underline{0.393} & \underline{0.431} & 0.441 & 0.451 & 0.832 & 0.654 & 0.616 & 0.558 & 0.491 & 0.491 & 0.467 & 0.481 & 3.510 & 1.571 \\
        \midrule
        \multirow{4}{*}{\rotatebox{90}{ETTm1}}
        & 96 & \textbf{0.292} & \textbf{0.344} & \underline{0.295} & \underline{0.344} & 0.335 & 0.374 & 0.323 & 0.369 & 0.303 & 0.347 & 0.354 & 0.406 & 0.487 & 0.470 & 0.734 & 0.631 \\
        & 192 & \textbf{0.333} & \underline{0.369} & 0.338 & 0.373 & 0.387 & 0.399 & 0.356 & 0.394 & \underline{0.336} & \textbf{0.366} & 0.391 & 0.424 & 0.573 & 0.507 & 0.756 & 0.642 \\
        & 336 & \textbf{0.367} & \underline{0.390} & \underline{0.370} & 0.393 & 0.421 & 0.426 & 0.407 & 0.427 & 0.372 & \textbf{0.389} & 0.433 & 0.453 & 0.550 & 0.501 & 1.088 & 0.806 \\
        & 720 & \underline{0.421} & \underline{0.422} & \textbf{0.418} & \textbf{0.422} & 0.483 & 0.460 & 0.479 & 0.475 & 0.429 & 0.425 & 0.491 & 0.479 & 0.533 & 0.507 & 1.032 & 0.786 \\
        \midrule
        \multirow{4}{*}{\rotatebox{90}{ETTm2}}
        & 96 & \textbf{0.164} & \textbf{0.253} & \underline{0.165} & \underline{0.255} & 0.184 & 0.265 & 0.185 & 0.280 & 0.168 & 0.262 & 0.188 & 0.281 & 0.249 & 0.323 & 0.347 & 0.442 \\
        & 192 & \textbf{0.222} & \textbf{0.293} & \underline{0.222} & \underline{0.294} & 0.259 & 0.309 & 0.246 & 0.319 & 0.237 & 0.318 & 0.255 & 0.326 & 0.281 & 0.338 & 0.843 & 0.717 \\
        & 336 & \textbf{0.274} & \textbf{0.327} & \underline{0.275} & \underline{0.330} & 0.314 & 0.346 & 0.329 & 0.383 & 0.294 & 0.356 & 0.323 & 0.367 & 0.339 & 0.376 & 1.450 & 0.934 \\
        & 720 & \textbf{0.364} & \textbf{0.381} & \underline{0.367} & \underline{0.385} & 0.424 & 0.408 & 0.468 & 0.468 & 0.403 & 0.423 & 0.426 & 0.421 & 0.416 & 0.415 & 3.548 & 1.431 \\
        \midrule
        \multirow{4}{*}{\rotatebox{90}{Weather}}
        & 96 & \underline{0.150} & \underline{0.198} & \textbf{0.149} & \textbf{0.197} & 0.171 & 0.222 & 0.171 & 0.240 & 0.177 & 0.239 & 0.228 & 0.313 & 0.290 & 0.358 & 0.389 & 0.447 \\
        & 192 & \textbf{0.192} & \textbf{0.238} & \underline{0.194} & \underline{0.240} & 0.223 & 0.263 & 0.212 & 0.275 & 0.219 & 0.279 & 0.276 & 0.342 & 0.302 & 0.363 & 0.461 & 0.470 \\
        & 336 & \textbf{0.244} & \textbf{0.278} & \underline{0.246} & \underline{0.281} & 0.277 & 0.301 & 0.273 & 0.333 & 0.265 & 0.317 & 0.326 & 0.368 & 0.358 & 0.389 & 0.540 & 0.504 \\
        & 720 & \textbf{0.323} & \textbf{0.332} & \underline{0.324} & \underline{0.335} & 0.355 & 0.351 & 0.347 & 0.385 & 0.332 & 0.374 & 0.399 & 0.407 & 0.432 & 0.436 & 1.258 & 0.815 \\
        \midrule
        \multirow{4}{*}{\rotatebox{90}{Electricity}}
        & 96 & \underline{0.136} & \underline{0.231} & \textbf{0.130} & \textbf{0.224} & 0.169 & 0.273 & 0.161 & 0.267 & 0.140 & 0.238 & 0.192 & 0.306 & 0.213 & 0.328 & 0.331 & 0.416 \\
        & 192 & \underline{0.152} & \underline{0.245} & \textbf{0.149} & \textbf{0.242} & 0.189 & 0.291 & 0.177 & 0.286 & 0.154 & 0.251 & 0.205 & 0.319 & 0.244 & 0.350 & 0.360 & 0.441 \\
        & 336 & \underline{0.168} & \underline{0.262} & \textbf{0.166} & \textbf{0.260} & 0.201 & 0.303 & 0.199 & 0.304 & 0.170 & 0.268 & 0.217 & 0.331 & 0.247 & 0.349 & 0.361 & 0.442 \\
        & 720 & \underline{0.206} & \textbf{0.294} & 0.211 & \underline{0.301} & 0.221 & 0.318 & 0.215 & 0.323 & \textbf{0.204} & 0.302 & 0.245 & 0.353 & 0.321 & 0.402 & 0.482 & 0.519 \\
        \midrule
        \multirow{4}{*}{\rotatebox{90}{Traffic}}
        & 96 & \underline{0.387} & \underline{0.265} & \textbf{0.366} & \textbf{0.249} & 0.589 & 0.318 & 0.511 & 0.305 & 0.411 & 0.284 & 0.582 & 0.367 & 0.672 & 0.426 & 0.768 & 0.435 \\
        & 192 & \underline{0.400} & \underline{0.269} & \textbf{0.386} & \textbf{0.259} & 0.616 & 0.328 & 0.534 & 0.308 & 0.424 & 0.289 & 0.606 & 0.377 & 0.659 & 0.414 & 0.815 & 0.462 \\
        & 336 & \underline{0.411} & \underline{0.275} & \textbf{0.397} & \textbf{0.264} & 0.643 & 0.347 & 0.534 & 0.311 & 0.437 & 0.298 & 0.623 & 0.385 & 0.619 & 0.386 & 1.015 & 0.576 \\
        & 720 & \underline{0.439} & \underline{0.291} & \textbf{0.431} & \textbf{0.283} & 0.669 & 0.358 & 0.565 & 0.323 & 0.467 & 0.317 & 0.639 & 0.391 & 0.680 & 0.420 & 1.462 & 0.798 \\
        \midrule
        \multirow{4}{*}{\rotatebox{90}{ILI}}
        & 24 & \textbf{1.840} & \textbf{0.840} & 2.073 & 0.923 & \underline{1.851} & \underline{0.896} & 2.649 & 1.124 & 2.253 & 1.049 & 3.560 & 1.353 & 3.559 & 1.344 & 5.349 & 1.574 \\
        & 36 & \textbf{1.807} & \textbf{0.862} & \underline{2.100} & \underline{0.946} & 2.792 & 1.010 & 2.740 & 1.137 & 2.252 & 1.059 & 3.450 & 1.297 & 3.880 & 1.383 & 5.034 & 1.534 \\
        & 48 & \underline{2.052} & 0.942 & \textbf{1.852} & \underline{0.936} & 2.491 & \textbf{0.929} & 2.830 & 1.140 & 2.280 & 1.074 & 3.161 & 1.226 & 3.247 & 1.240 & 5.130 & 1.554 \\
        & 60 & \textbf{1.764} & \underline{0.886} & 2.295 & 1.038 & \underline{1.909} & \textbf{0.876} & 2.838 & 1.141 & 2.392 & 1.106 & 3.239 & 1.257 & 3.164 & 1.224 & 5.380 & 1.601 \\
        \bottomrule
    \end{tabular}}
    \label{tab:main_LTF}
\end{table*}
\def\arraystretch{1.3}
\begin{table*}[ht]
    \footnotesize
    \caption{Univariate short-term forecasting results with MultiResFormer. The best results are in \textbf{bold} and the second best are \underline{underlined}. Baseline results are taken from \cite{Wu2023TimesNet}.}
    \centering
    \begin{adjustbox}{width=0.85\linewidth}
    \begin{tabular}{c|c|C{1.41cm}|C{1.41cm}|C{1.41cm}|C{1.41cm}|C{1.41cm}|C{1.41cm}|C{1.41cm}|C{1.41cm}}
        \toprule
        \multicolumn{2}{c|}{Models} & MultiRes. & TimesNet & N-HiTS & N-BEATS & DLinear & FED. & Auto. & In. \\
        \midrule
        \multirow{3}{*}{\rotatebox{90}{Yearly}}
        & SMAPE & \textbf{2.975} & \underline{2.996} & 3.045 & 3.043 & 4.283 & 3.048 & 3.134 & 3.418 \\
        & MASE & \textbf{2.975} & \underline{2.996} & 3.045 & 3.043 & 4.283 & 3.048 & 3.134 & 3.418 \\
        & OWA & \textbf{0.780} & \underline{0.786} & 0.793 & 0.794 & 1.058 & 0.803 & 0.822 & 0.881 \\
        \midrule
        \multirow{3}{*}{\rotatebox{90}{Quarterly}}
        & SMAPE & \textbf{10.026} & \underline{10.100} & 10.202 & 10.124 & 12.145 & 10.792 & 11.338 & 11.360 \\
        & MASE & \textbf{1.169} & \underline{1.182} & 1.194 & \textbf{1.169} & 1.520 & 1.283 & 1.365 & 1.401 \\
        & OWA & \textbf{0.882} & 0.890 & 0.899 & \underline{0.886} & 1.106 & 0.958 & 1.012 & 1.027 \\
        \midrule
        \multirow{3}{*}{\rotatebox{90}{Monthly}}
        & SMAPE & 12.854 & \textbf{12.670} & 12.791 & \underline{12.677} & 13.514 & 14.260 & 13.958 & 14.062 \\
        & MASE & 0.960 & \textbf{0.933} & 0.969 & \underline{0.937} & 1.037 & 1.102 & 1.103 & 1.141 \\
        & OWA & 0.897 & \textbf{0.878} & 0.899 & \underline{0.880} & 0.956 & 1.012 & 1.002 & 1.024 \\
        \midrule
        \multirow{3}{*}{\rotatebox{90}{Others}}
        & SMAPE & \textbf{4.612} & \underline{4.891} & 5.061 & 4.925 & 6.709 & 4.954 & 5.485 & 24.460 \\
        & MASE & \textbf{3.140} & 3.302 & \underline{3.216} & 3.391 & 4.953 & 3.264 & 3.865 & 20.960 \\
        & OWA & \textbf{0.980} & \underline{1.035} & 1.040 & 1.053 & 1.487 & 1.036 & 1.187 & 5.879 \\
        \bottomrule
    \end{tabular}
    \end{adjustbox}
    \label{tab:main_STF}
\end{table*}


\end{document}